# Can Unsupervised Methods Outperform Supervised Deep Learning When Ground Truth Is Sparse? A Case Study of Bronchovascular Bundle Segmentation in Low-Dose CT

Anna Mrukwa[1§], Marek Socha[1§], Aleksandra Suwalska[1], Agata Durawa[2], Malgorzata Jelitto[3], Katarzyna Dziadziuszko[3], Edyta Szurowska[3], Pawel Bożek[4], Michal Marczyk[1,5], Witold Rzyman[2], Rafal Dziadziuszko[6], and Joanna Polańska[1*]

[1] Department of Data Science and Engineering, Silesian University of Technology, Gliwice, Poland
[2] Department of Thoracic Surgery, Medical University of Gdańsk, Gdańsk, Poland
[3]2nd Division of Radiology, Medical University of Gdansk, Gdansk, Poland
[4] Department of Radiology and Radiodiagnostics, Medical University of Silesia, Katowice, Poland
[5] Department of Breast Medical Oncology, Yale School of Medicine, New Haven, CT, USA
[6]Department of Oncology and Radiotherapy, Medical University of Gdańsk, Gdańsk, Poland

§ These authors contributed equally
* Correspondence: joanna.polanska@polsl.pl, Department of Data Science and Engineering, Faculty of Automatic Control, Electronics and Computer Science, Akademicka 16, 44-100 Gliwice, Poland

## Acknowledgments

### Guarantor statement

J.P. takes full responsibility for the contents of the manuscript, including the data and analysis.

### Statement

During the preparation of this work the authors used Perplexity to improve the language and the manuscript style, as well as DeepL for translation purposes. After using this service, the authors reviewed and edited the content as needed and take full responsibility for the content of the publication

### Disclosures

This work has been supported by the Silesian University of Technology grant for maintaining and developing research potential [grant no. 02/070/BKM_26/0085] which supported the research activities over the course of the project and will also cover publication costs, and by Pomeranian Interdisciplinary Centre of Digital Medicine [grant no. 2023/ABM/02/00018]. The LDCT examinations carried out as part of the Pilot Pomeranian Lung Cancer Screening Program in years 2008-2011 were co-financed by the Financial Mechanism of the European Economic Area and funds from the Pomeranian Subsidy.

Part of this research was submitted to the international conference European Molecular Imaging Meeting 2026.

## Abbreviations

CAD – Computer-Aided Diagnosis

CT – Computed Tomography

DLCS – Duke Lung Cancer Screening

HARS – high attenuation rough segmentation

HU – Hounsfield Unit

LARS – low attenuation rough segmentation

LDCT – low dose Computed Tomography

## Abstract

### *Background*

Lung cancer remains the deadliest of the cancers worldwide due to the late diagnosis. To prevent its development and enable treatment, it must be diagnosed early, during screening. However, increasing number of patients and limited number of radiologists, results in prolonged diagnostic waiting time. Additionally, nodules visibility in very early stage lung cancer are reduced by blood vessels and airway walls as nodules are supplied by them. The task-specific analysis of the bronchovascular bundle is thus important for efficient nodule detection and its removal will increase the diagnostic potential for lung cancer.

### *Materials and Methods*

To assess the efficacy of the proposed methods, series from widely utilized LDCT datasets were used, such as DLCS and Pilot Pomeranian Lung Cancer Screening Program collections. The proposed bronchovascular bundle segmentation pipeline (RONALD) is based on the Computed Tomography images and returns binary masks of vessels and bronchi situated in the lung parenchyma. The method has preprocessing stage, where segmentation of lungs, lobes and mediastinum is carried out, then divides into separate vessels and bronchial tree segmentation.

### *Results*

Created pipeline can segment the bronchovascular bundle in the low dose computer tomography scans, with the nodule retaining improvement in comparison with other segmentation methods from 93.98% and 90.36% to 100% (DLCS) and 83.16%, 62.36% to 99.92% (Pomeranian).

### *Conclusion*

The resulting segmentations enable better lung nodule detection in the very early stages of lung cancer.

## Introduction

Lung cancer screening programs are a cost-effective approach for early diagnosis and have been shown to improve patient survival[1–3]. However, the large number of acquired CT scans, combined with the limited availability of radiologists, results in long waiting times for diagnostic decisions, during which patients may remain untreated. This creates a strong need for computer-aided diagnosis (CAD) systems to support clinical decision-making. Very early stage cancerous nodules are typically small and often connected to the bronchovascular bundle, which makes them difficult to detect. Even standard CAD systems based on unprocessed lung CT images frequently fail in such cases, as nodules attached to vessels or bronchi are hard to distinguish from surrounding structures. This is consistent with recent LDCT detection work showing that vessels and bronchi are major confounders for pulmonary nodule detection, underscoring the importance of bronchovascular anatomy in early lung cancer detection[4]. Therefore, in addition to conventional parenchymal analysis, the bronchovascular tree should be explicitly analyzed. For this reason, lung nodule-aware segmentation of the bronchovascular tree, as proposed in our method, is a crucial component of effective very early stage lung cancer detection.

Low dose computed tomography (LDCT)[5] is a common diagnostic tool employed in the screening programs to facilitate early detection of lung cancer, while not using the full radiation dose for patients with no changes. The dosage standard depends on the country conducting the screening programme, so the low dose scans may differ in the terms of noise and artifacts – in the European Union, the suggested dose should be less than 1mSv[6], whereas in the United States the dose is around 2mSv[7]. Because of that, the creation of the system for the nodule detection is hindered, as the data differs greatly depending on its source. Thus, the models trained on the popular NLST dataset[8] may not be applicable for the studies performed using ultra-low dose computer tomography, as they are closer to standard dose in some countries.

In the context of lung diseases, the analysis of the bronchovascular bundle's appearance is paramount for accurate diagnosis of lower airways diseases. Moreover, the process of bronchovascular bundle segmentation generates indirect benefits, as the bundle contains the vessels that supply the lung nodules. Because of that, the presence of lung nodules in proximity to the visible parts of the bronchovascular bundle is a notable observation. The identification of these lesions is complicated by their connection and similar voxel intensity to the bundle, which makes them more difficult to identify, whether by radiologists or CAD systems.

Segmenting the bronchovascular bundle in LDCT lung scans poses considerable challenges due to increased noise, low contrast, and anatomical complexity. In the context of airway segmentation, conventional methodologies such as region growing[9] and fuzzy connectivity[10,11] have been shown to enhance thicker airway branches. However, these approaches frequently encounter leakage into the surrounding lung parenchyma, a phenomenon that is particularly pronounced in LDCT scans, where the airway wall is thin and image noise is high. The leakage causes the segmentation to incorrectly include non-airway regions. Frequently, the algorithms need to be manually tuned to achieve a balance between sensitivity and specificity. Furthermore, the process of segmentation often terminates at an earlier stage in peripheral bronchioles due to diminished contrast. The employment of hybrid methods, such as the freeze-and-grow technique, has been demonstrated to assist in mitigating leakage and enhancing the accuracy of branch detection[11,12]. However, these methods encounter challenges in scenarios involving severely compromised lungs and the presence of motion artifacts[12,13]. Deep learning approaches, particularly U-Net variants with specialized loss functions, have been shown to improve the detection of small bronchioles and maintain airway continuity[13–15]. However, these approaches are limited by class imbalance and the scarcity of high-quality annotated LDCT data for training[15,16]. This limitation can impede generalization and accuracy in peripheral airway segmentation.

For the vessel segmentation, classical vesselness filters based on Hessian matrix analysis have been shown to enhance tubular structures[17,18]. However, these filters encounter challenges in accurately segmenting small vessels and noisy LDCT images, as well in the appearance of the nodules connected to the vessels. Thus, the extraction of vascular trees is often facilitated by region growing and morphological operations[19].

However, these processes are susceptible to variations in parameter selection and the presence of image artifacts. The employment of graph-cut and rule-based methodologies has been demonstrated to enhance the delineation of boundaries[20,21]. Unfortunately, the tuning process requires special attention to choose parameters depending on the data source. Recent advancements in deep learning models, incorporating multi-scale feature fusion, have demonstrated considerable potential in accurately segmenting arteries and veins in LDCT scans, despite the presence of noise[22,23]. Once again, the efficacy of these models is dependent on the availability of large, annotated datasets, a limitation that is particularly pronounced for LDCT scans. This constraint impacts the models' robustness and clinical applicability. System adapted to the LDCT scans nature is needed.

The newest methods combine the airways and vessels segmentation in one deep learning pipeline. AirRC[24] provides expert 3D annotations of pulmonary arteries, veins, airway lumen and airway wall for 254 scans from the LUNA16 benchmark[25], alongside training scripts of nnUNet[26] models that accurately delineate the bronchovascular tree and capture fine peripheral branches. However, its moderate cohort size and absence of dedicated LDCT screening data, constrain the direct transfer of such models to noisy low-dose populations. TotalSegmentator[27,28], in turn, offers a robust, off-the-shelf nnUNet–based pipeline that segments more than one hundred anatomical structures, including lung vessels and central airways, on routine clinical CT and its bronchovascular labels have also been further refined using AirRC annotations, providing a strong starting point for bundle modelling. Yet, because of the same training dataset, its performance in lung cancer screening may be limited. All the methods are described in the Supplementary Tables 1, 2 and 3.

This study introduces a novel method (RONALD) for nodule-aware segmentation of the bronchovascular tree in low dose computed tomography (LDCT) scans, designed to enhance the visibility of pulmonary nodules. It hypothesis that the selective removal of bronchovascular structures enhances the lung nodules visibility and reduces the anatomical complexity in the regions of diagnostic interest, thereby enhancing the visibility of pulmonary nodules, including those adjacent to vessels or bronchi, which represent the most challenging cases for both radiologists and CAD systems. Given that undetected nodules at the screening stage are directly associated with delayed diagnosis and worse patient survival, nodule visibility is considered a clinically consequential outcome. The proposed method is designed to be integrated into existing radiological workflows and CAD pipelines as an upstream preprocessing step, operating on standard LDCT input without requiring manual interaction.

## Study design and methods

In this work, two datasets were retrospectively used to test the algorithm performance on different cohorts and resolutions: the Pomeranian Pilot Lung Cancer Screening Program[29,30], which was approved by the Bioethics Committee for Scientific Research at the Medical University of Gdańsk (approval no. NKEBN/109/2009), and written informed consent was obtained from all participants. The Duke Lung Cancer Screening (DLCS)[31] cohort was collected as part of a HIPAA-compliant retrospective study approved by the Duke University Health System Institutional Review Board (protocol Pro00108863), with a waiver of informed consent. Both datasets consist of thorax low dose computed tomography series and originate from the lung cancer screening programs. The chosen series included annotated coordinates of malignant and benign nodules used to assess the quality of the segmentations (Table 1).

Pilot Pomeranian Lung Cancer Screening Program (Pomeranian)[29,30] conducted between 2009 and 2011 consisted of 2002 patients, who underwent low-dose CT screening as a part of an early lung cancer detection initiative. The patients were between 50 and 75 years old and had been or had smoked for at least 20 years. The main objective of the program was early detection of lung cancer cases, which could be treated with surgical operation. Patients did not present any symptoms of the disease during the trial and were considered cancer free on the date of undergoing the screening. The subset of 1201 series was used as the main benchmark for evaluating the algorithm's performance due to the high number of annotated nodules locations available per series.

The Duke Lung Cancer Screening (DLCS)[31] dataset includes 1613 patients who underwent low-dose CT (LDCT) scans as part of lung cancer screening within the Duke Health system in 2015-2021. The patients are aged 55-80, with significant smoking histories. The dataset includes nodules annotated with 3D bounding boxes and classified according to the Lung-RADS lexicon, with lung cancer outcomes recorded. Dataset reflects real-world clinical screening practices, capturing a population with diverse sociodemographic backgrounds and comorbidities. This publicly available dataset supports research in lung cancer risk classification and diagnostic model development and can be accessed via Zenodo.

**Table 1:** Summary table of used datasets.

| Dataset | Average resolution | Number of series | Average voxel size | Nodules | |
|---|---|---|---|---|---|
| | | | | Malignant | Benign |
| Pomeranian | 512x512x288 | 1201 | 0.89x0.89x1.26 | 1271 | 3927 |
| DLCS | 512x512x536 | 1613 | 0.7x0.7x0.625 | 264 | 1538 |

The proposed algorithm (RONALD) consists of a series of steps, starting with pre-processing stage where the lungs were segmented. Based on the anatomical structure the mediastinum was identified. The algorithm then branches into two independent processes: one dedicated to the tracheobronchial tree segmentation, including both airway lumen and walls, and the other focused on pulmonary vessels. The diagram presenting the algorithm's pipeline is shown in Figure 1.

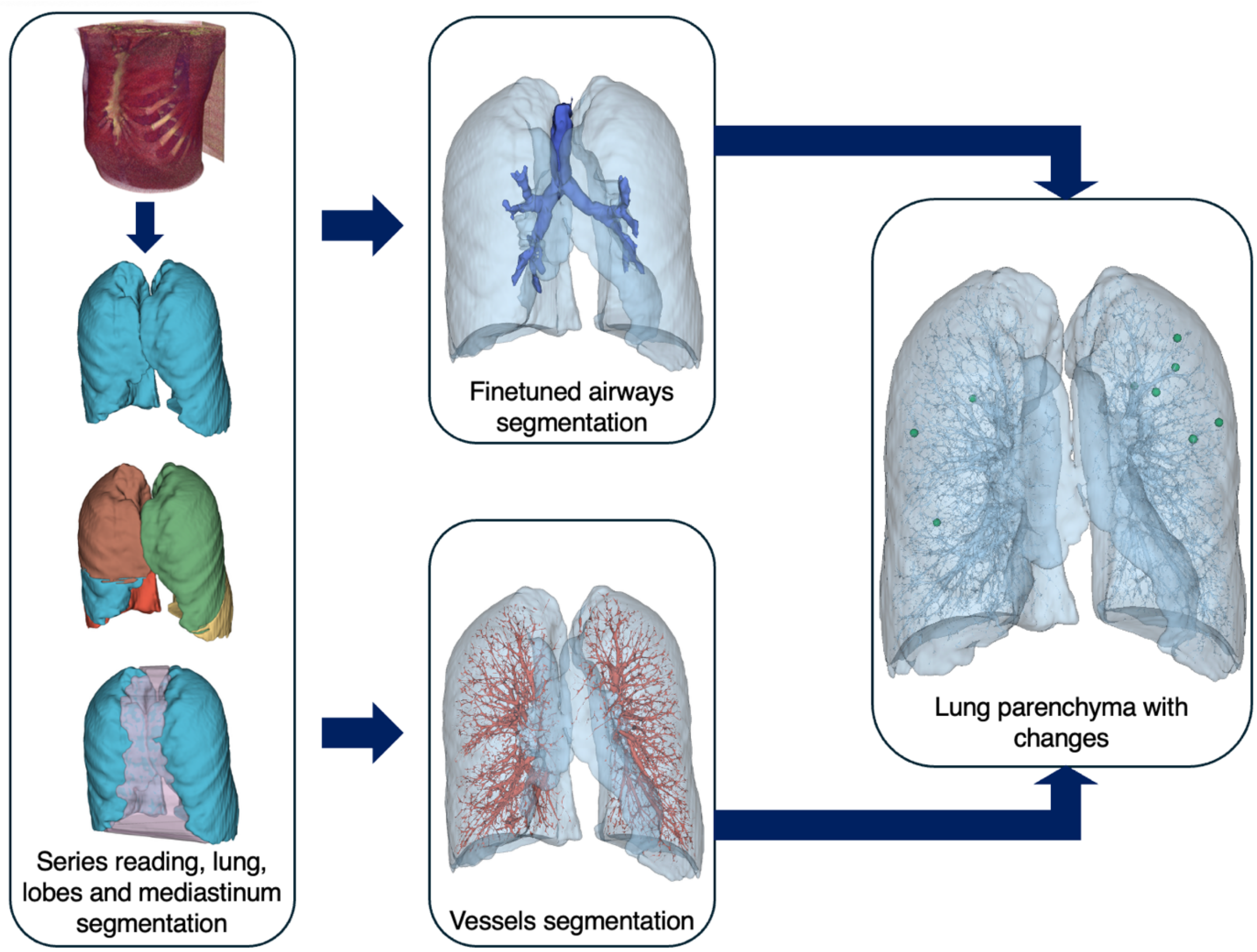

**Figure 1.** Processing pipeline for bronchovascular bundle modelling.

## *Lung and lobes segmentation*

Lung segmentation was performed automatically using TotalSegmentator (version 2.13.0), a pretrained deep learning–based medical image segmentation framework. For the present analysis, only the whole-lung masks generated by TotalSegmentator were used. The goal was to assign each pixel in the image to either the background or lung tissue class by creating common binary mask from the partial lobe masks. The masks are shown in the Supplementary Figure 1.

### *Mediastinum segmentation*

Mediastinal segmentation was required for the segmentations of trachea, as well as the vessels. It was extracted by subtracting lung mask from its convex hull. The resulting mask was cleaned up from residual noise by the morphological opening operation. Then, the biggest object was chosen as the mediastinum, by selecting the largest fully connected component. The mediastinum segmentation results are depicted in the Supplementary Figure 2.

### *Trachea segmentation*

Tracheal mask serves as the foundation for the bronchial modelling procedure and was used to estimate the air values in the scan. According to the literature, the theoretical median values of Hounsfield Units (HU) are between -1024 HU and -920 HU[32] for the air. Thresholding within this range typically results in multiple disconnected regions in three-dimensional space. The trachea was then identified as the largest and most centrally located of these regions.

### *Rough tracheobronchial tree segmentation*

Firstly, the rough mask of high and low attenuation regions was created via Gaussian Mixture Model[33]. Low attenuation rough segmentation (LARS) represents air and diseases such as emphysema, while high attenuation rough segmentation (HARS) represents objects like solids, walls, fluids, consolidations, nodules, and vessels. Then, using LARS, speed map was created. Low intensity regions were processed by calculating their gradient magnitude, which was further combined with the tracheobronchial tree enhanced by the Sato filter. Then, based on the trachea segmentation, the seed points were chosen as the air values inside the tracheobronchial tree. Such prepared input was fed into the fast-marching algorithm. The result was binarized and compared with the trachea segmentation. The component exhibiting the largest number of overlapping voxels with the trachea was then selected as the initial airway mask.

### *Finetuned tracheobronchial tree segmentation*

Due to the nature of the fast-marching algorithm, previously segmented regions could include over-segmentations. Thus, the walls were obtained through a combination of airway dilation and the mask created by one of the HARS from the Gaussian mixture model.

The closer to the terminal bronchioli, the more the boundary between the bronchial tree and the surrounding parenchyma becomes indistinguishable. Consequently, the walls segmentation was further finetuned by approximating the bronchi with the mathematical model under the assumption that the branches are tube-like structures. Because of that, the skeleton of the tree was found using Lee's method[34], similarly like in [13]. If the method found too many branches due to the oversegmentation by the fast marching, the skeletonization was instead performed using the Kimimaro approach[35]. The tree was divided into segments based on its skeleton and these segments were approximated by the cylinders. The segments were chosen adaptively using the Visvalingam–Whyatt algorithm[36], so that the bends in the branches were properly modelled. Then, to connect the branches, convex hull and adaptive closing of the branches with the element depending on the branch size were performed. This reduced the number of leakages and removed the nodules accumulated on the bronchi walls.

## Vessel segmentation

To complete segmentation of the bronchovascular bundle, the blood vessels in the lungs were segmented using Frangi filter. To reduce the artifact amount, using the connectivity criterion, the vessels connected to the mediastinum were retained. Moreover, the vessels should be connected inside the lobes and not between them due to the connectivity. As many of the malignant changes are supplied by the blood vessels, the blob-like objects[37] were filtered out from the vessels mask. Whole bronchovascular bundle segmentation is shown in the Figure 2.

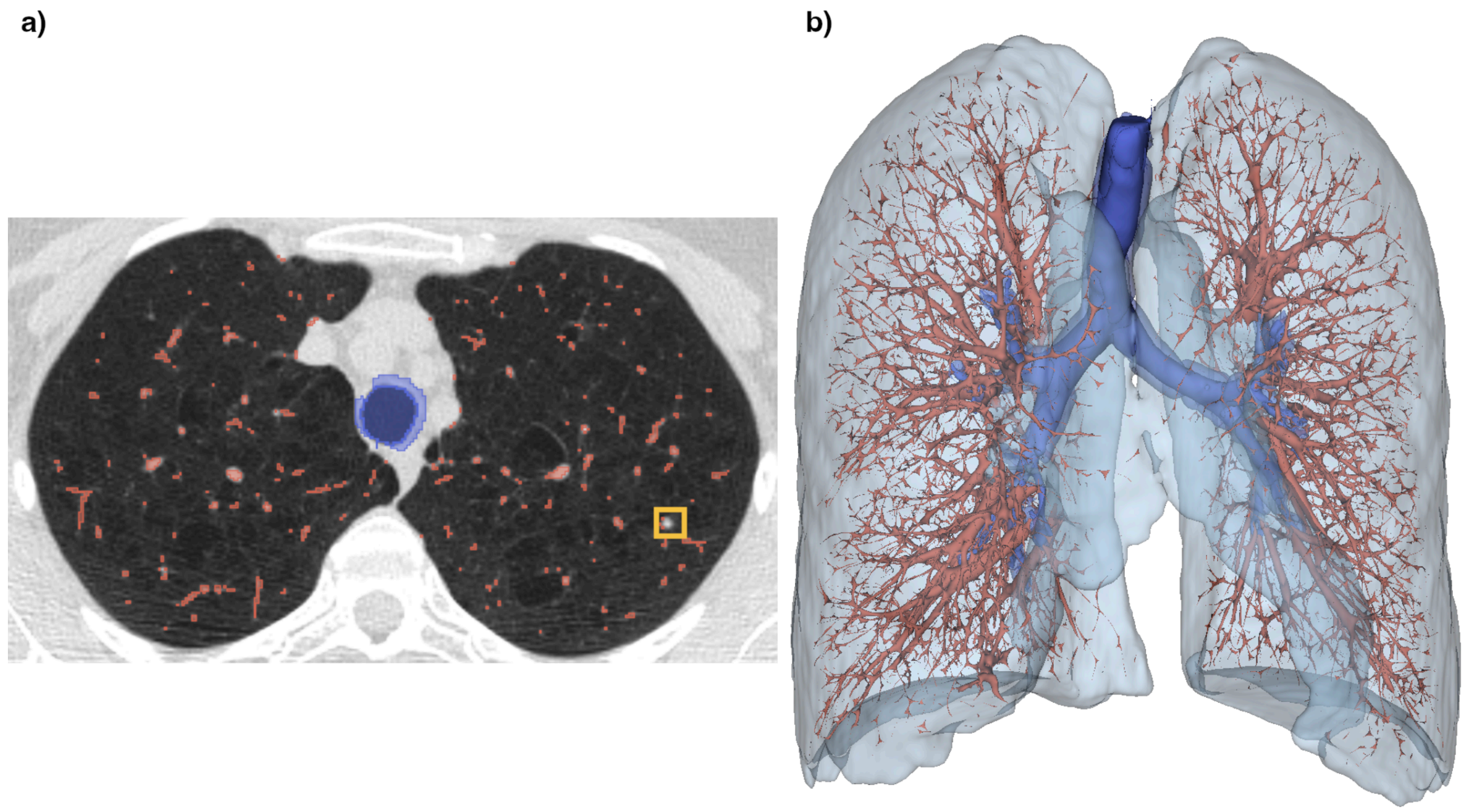

**Figure 2.** Results of RONALD's bronchovascular bundle segmentation. The tracheobronchial walls are slightly lighter to accentuate the airflow. The marked malignant nodule is excluded from the structures.

## Evaluation

To evaluate if the segmentation of the bronchovascular tree improved the visibility of pulmonary nodules, a comparison to two other bronchovascular bundle segmentation methods was made: AirRC's nnUNet and TotalSegmentator. The comparison was based on analysing how many nodules remained within the lung parenchyma in each mask by checking if the provided annotation coordinates were still inside the masks. In particular, per patient detected ratio was calculated as the number of detected nodules divided by the total number of nodules found by the radiologists. Exemplary masks are shown in the Figures 3 and 4, respectively.

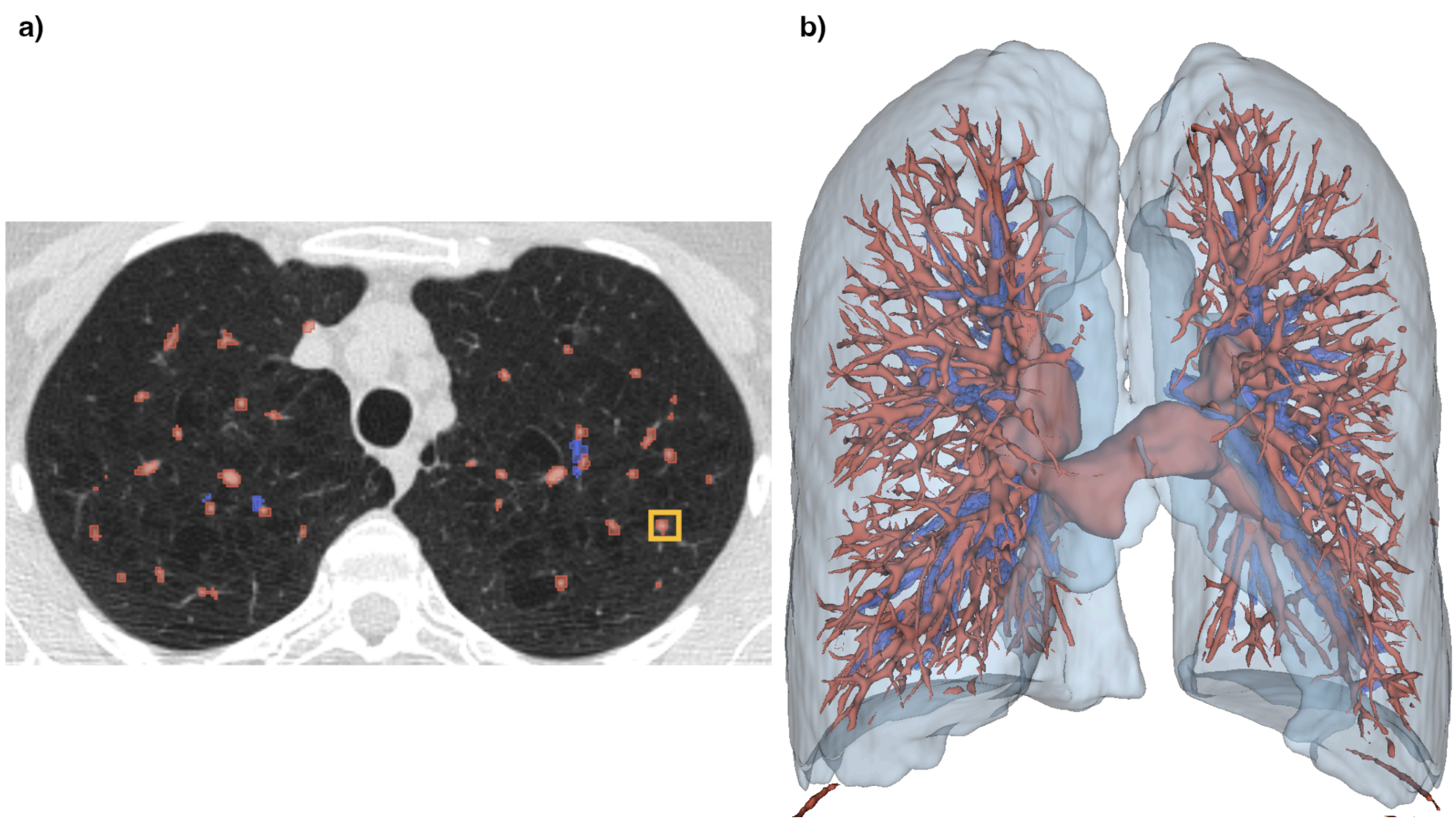

**Figure 3.** Results of AirRC's bronchovascular bundle segmentation. The marked malignant nodule is inside the structures.

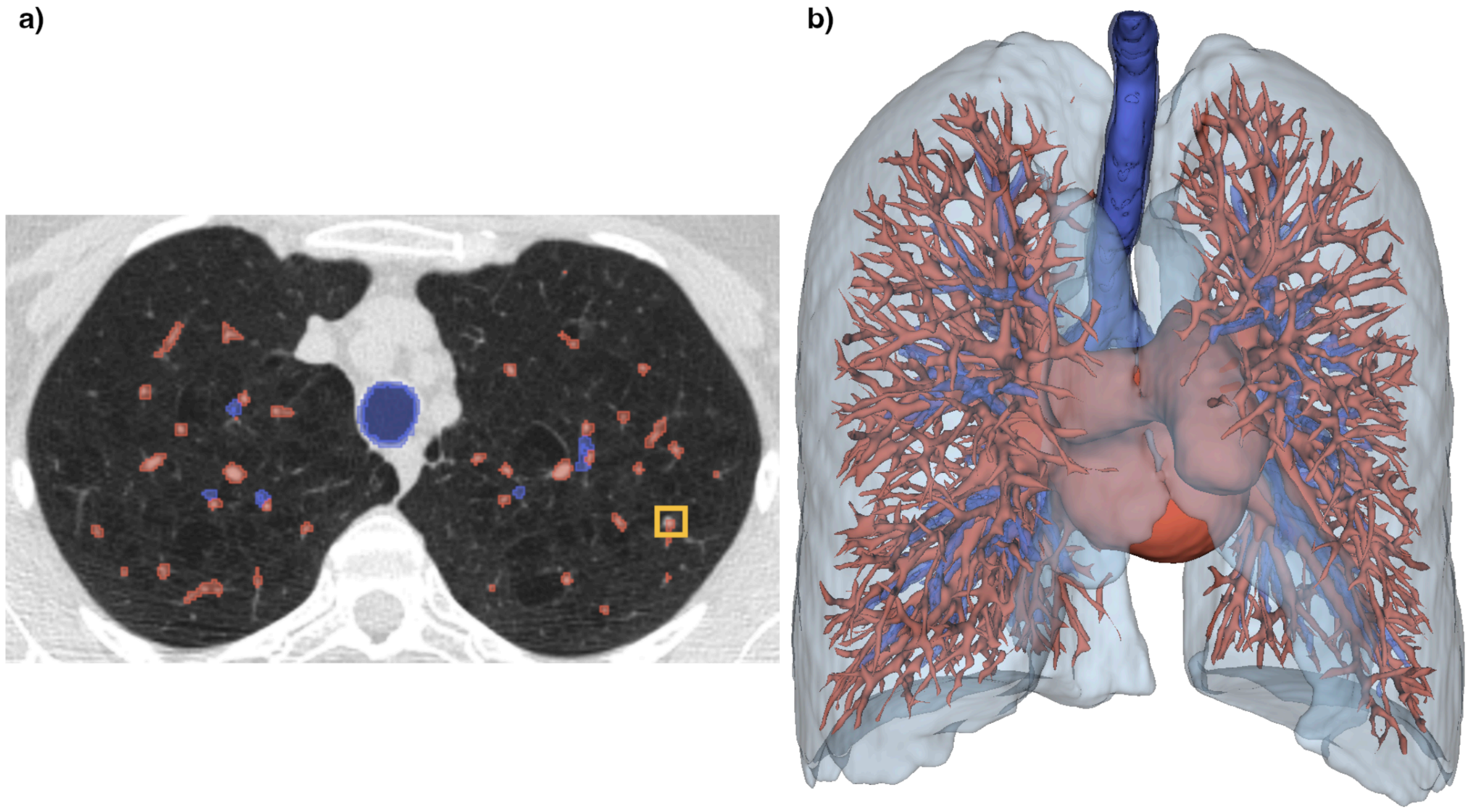

**Figure 4**. Results of TotalSegmentator's bronchovascular bundle segmentation. The marked malignant nodule is inside the structures.

## Results

RONALD preserved almost all nodules in both datasets, outperforming both reference methods. For the AirRC segmentations, out of 1271 malignant nodules used from the Pomeranian dataset, 1057 (83.16%) remained in parenchyma after filtering, while the rest of lung nodules were lost. In case of the DLCS dataset, the method leaves 156 (93.98%) nodules out of 166 used for validation.

As for the TotalSegmentator model, only 792 nodules (62.36%) remained in the parenchyma for Pomeranian dataset. For DLCS dataset, the algorithm retained 150 (90.36%) nodules.

In turn, after the removal of the bronchovascular bundle in Pomeranian dataset all but one were still in the parenchyma (99.92%). For DLCS, segmentation and the removal of bronchovascular bundle result in retaining all nodules (100%). The results are presented in the Table 2.

Per patient detected ratio was calculated for each dataset and approach and then methods were compared using Wilcoxon signed-rank test. Because the matched-pairs rank-biserial correlation was uniformly saturated at 1.0 in these comparisons, effect magnitude was summarized using Cohen's d for the paired differences. The difference between the supervised methods and RONALD is significant for both datasets, with the p-value for the Pomeranian dataset smaller than $1*10^{-10}$ for AirRC, with the effect size 0.542 and the p-value for the DLCS smaller than $1*10^{-10}$ and the effect size 0.3043. For TotalSegmentator, the difference is even bigger, with p-values for both datasets smaller than $1*10^{-10}$, effect size for Pomeranian 0.970, while for DLCS 0.369.

**Table 2.** Comparison of methods for the separation of vessels and airway walls from nodules. AirRC and TotalSegmentator retain less objects than RONALD.

| Dataset | Validation nodules | Retained after AirRC | Retained after TotalSegmentator | Retained after RONALD |
|---|---|---|---|---|
| Pomeranian | 1271 | 1057 | 792 | 1270 |
| DLCS | 166 | 156 | 150 | 166 |

## Discussion

The single nodule removed in the Pomeranian dataset resulted from leakage of the bronchial tree into the low attenuation regions in their peripheral. The patient had other coexisting ailments, such as emphysema, making the bronchovascular bundle less distinct. The removal of the one change did not make the diagnosis impossible, as 15 distinct nodules were annotated.

The higher apparent lesion retention in DLCS for AirRC and TotalSegmentator may partly reflect acquisition differences rather than method behaviour alone. DLCS CTs are higher-dose than the Pomeranian scans, which would make lesion-adjacent positive-HU structure easier to preserve or recover under the outside-mask criterion.

When skeletonizing the rough bronchial tree segmentation with a significant number of changes and artifacts connected to it, the Lee's algorithm overcomplicates the skeleton structure. The trachea is no longer represented by single edge with two end nodes, but the artifacts connected to it are seen as additional end nodes, creating additional extensions from the main trachea axis. To overcome this and simplify the structure, the Kimimaro algorithm is used when the nodes number is an outlier based on the Pomeranian dataset values. This way, the skeleton structure is less complex, and modelling becomes faster and cleaner. The cutoff value was set based on the Pomeranian cohort statistics and applied without modifications to the DLCS dataset and while it gave satisfactory results for the DLCS dataset, it may be beneficial to adapt it based on the dataset characteristics. For the used datasets the data is similar enough to apply the same cutoff value, as shown in the Supplementary Figure 3.

The final smoothing by closing of the modeled branches is adaptively performed based on their size, with the branches narrower than half of the trachea having morphological element of size one, the rest two. This was estimated as the best threshold, giving smooth but not overgrown branches. The closing element is chosen based on the definition of the early-stage changes – the nodule is smaller than 3 cm, whereas the closing element has the diameter around 3.5 mm, depending on the voxel size.

Prepared bronchovascular bundle modelling algorithm creates a hierarchical structure of branches and nodes reflecting the nature of the tracheobronchial tree and vessels. This, in turn, can be used for the analysis of the patient anatomy and improve the diagnosis process. In particular, the system enables the analysis of the early-stage lung cancer nodules, even the ones adjacent to the bronchovascular bundle. The method successfully segments the bronchovascular bundle while retaining the nodules (99.92% and 100% for Pomeranian and DLCS, respectively), which has potential to improve the diagnostic performance for early-stage lung cancer and has the potential to improve the diagnostic capabilities for other diseases, such as sarcoidosis. It is the first such method focusing on the nodule detection, with other available segmentation methods not modelling the airway walls or not being adapted to low dose CT scans.

The algorithm can be easily integrated into radiological consoles and other pipelines as it is compatible with the common medical image data formats. The source code is available and free to use at https://github.com/ZAEDPolSl/RONALD.

## Conclusions

Task-specific, nodule-aware segmentation of the bronchovascular bundle can substantially reduce the anatomical complexity that obscures early-stage pulmonary nodules in LDCT screening. With RONALD, bronchovascular structures are correctly separated from adjacent nodules in 99.92% and 100% of cases in the Pomeranian and DLCS cohorts respectively, with both results found to be significantly superior to those achieved by AirRC and TotalSegmentator. As undetected nodules at the screening stage represent missed opportunities for curative intervention, even modest improvements in nodule visibility are considered to carry meaningful patient-level consequences. The algorithm is freely available, compatible with standard medical imaging formats, and is designed to be incorporated as an upstream component in larger CAD or radiological reporting pipelines.

## References

1. Lam S, Bai C, Baldwin DR, et al. Current and Future Perspectives on Computed Tomography Screening for Lung Cancer: A Roadmap From 2023 to 2027 From the International Association for the Study of Lung Cancer. *J Thorac Oncol*. 2024;19(1):36-51. doi:10.1016/j.jtho.2023.07.019

2. Stone E, Marshall H, Yang PC, Fong KM. Lessons Learned From International Lung Cancer Screening Trials; People at Risk Deserve Screening for Early Detection. *Respirology*. 2025;30(9):802-816. doi:10.1111/resp.70097

3. Yip R, Mulshine JL, Oudkerk M, et al. Current evidence of low-dose CT screening benefit. *Eur J Cancer*. 2025;225. doi:10.1016/j.ejca.2025.115570

4. Yan S, Chang Y. TAPNet: A Geometry-Aware Triaxial Projection Network for Efficient Volumetric Pulmonary Nodule Detection.

5. Naidich DP, Marshall CH, Gribbin C, Arams RS, McCauley DI. Low-dose CT of the lungs: preliminary observations. *Radiology*. 1990;175(3):729-731. doi:10.1148/radiology.175.3.2343122

6. Revel MP, Biederer J, Nair A, et al. ESR Essentials: lung cancer screening with low-dose CT—practice recommendations by the European Society of Thoracic Imaging. *Eur Radiol*. Published online August 23, 2025. doi:10.1007/s00330-025-11910-9

7. Mascalchi M, Sali L. Lung cancer screening with low dose CT and radiation harm-from prediction models to cancer incidence data. *Ann Transl Med*. 2017;5(17):360. doi:10.21037/atm.2017.06.41

8. Aberle DR, Adams AM, Berg CD, et al. Reduced Lung-Cancer Mortality with Low-Dose Computed Tomographic Screening. *N Engl J Med*. 2011;365(5):395-409. doi:10.1056/NEJMoa1102873

9. Wang X, Fang C, Xia Y, Feng D. Airway segmentation for low-contrast CT images from combined PET/CT scanners based on airway modelling and seed prediction. *Biomed Signal Process Control*. 2011;6(1):48-56. doi:10.1016/j.bspc.2010.05.002

10. Rizi FY, Ahmadian A, Rezaie N, Iranmanesh SA. Leakage suppression in human airway tree segmentation using shape optimization based on fuzzy connectivity method. *Int J Imaging Syst Technol*. 2013;23(1):71-84. doi:10.1002/ima.22040

11. Tschirren J, Hoffman EA, McLennan G, Sonka M. Intrathoracic Airway Trees: Segmentation and Airway Morphology Analysis from Low-Dose CT Scans. *IEEE Trans Med Imaging*. 2005;24(12):1529-1539. doi:10.1109/TMI.2005.857654

12. Charbonnier JP, Rikxoort EM van, Setio AAA, Schaefer-Prokop CM, Ginneken B van, Ciompi F. Improving airway segmentation in computed tomography using leak detection with convolutional networks. *Med Image Anal*. 2017;36:52-60. doi:10.1016/j.media.2016.11.001

13. Sa N, Ea H, Jc S, et al. A CT-Based Automated Algorithm for Airway Segmentation Using Freeze-and-Grow Propagation and Deep Learning. *IEEE Trans Med Imaging*. 2021;40(1). doi:10.1109/TMI.2020.3029013

14. Garcia-Uceda A, Selvan R, Saghir Z, Tiddens HAWM, de Bruijne M. Automatic airway segmentation from computed tomography using robust and efficient 3-D convolutional neural networks. *Sci Rep*. 2021;11(1):16001. doi:10.1038/s41598-021-95364-1

15. Yang B, Liao H, Huang X, et al. Multi-Stage Airway Segmentation in Lung CT Based on Multi-scale Nested Residual UNet. *arXiv*. Preprint posted online October 24, 2024:arXiv:2410.18456. doi:10.48550/arXiv.2410.18456

16. Nadeem SA, Comellas AP, Hoffman EA, Saha PK. Generalizability of a deep learning airway segmentation algorithm to a blinded low-dose CT dataset. In: *Medical Imaging 2021: Image Processing*. Vol 11596. SPIE; 2021:951-958. doi:10.1117/12.2580224

17. Jimenez-Carretero D, Santos A, Kerkstra S, Rudyanto RD, Ledesma-Carbayo MJ. 3D Frangi-based lung vessel enhancement filter penalizing airways. In: *2013 IEEE 10th International Symposium on Biomedical Imaging*. 2013:926-929. doi:10.1109/ISBI.2013.6556627

18. Jin J, Yang L, Zhang X, Ding M. Vascular Tree Segmentation in Medical Images Using Hessian-Based Multiscale Filtering and Level Set Method. *Comput Math Methods Med*. 2013;2013(1):502013. doi:10.1155/2013/502013

19. Tan W, Yuan Y, Chen A, Mao L, Ke Y, Lv X. An Approach for Pulmonary Vascular Extraction from Chest CT Images. *J Healthc Eng*. 2019;2019:9712970. doi:10.1155/2019/9712970

20. Fabijańska A. Segmentation of pulmonary vascular tree from 3D CT thorax scans. *Biocybern Biomed Eng*. 2015;35(2):106-119. doi:10.1016/j.bbe.2014.07.001

21. Zhai Z, Staring M, Stoel BC. Lung vessel segmentation in CT images using graph-cuts. In: *Medical Imaging 2016: Image Processing*. Vol 9784. SPIE; 2016:699-706. doi:10.1117/12.2216827

22. Mank QJ, Thabit A, Maat APWM, et al. Artificial intelligence-based pulmonary vessel segmentation: an opportunity for automated three-dimensional planning of lung segmentectomy. *Interdiscip Cardiovasc Thorac Surg*. 2025;40(5):ivaf101. doi:10.1093/icvts/ivaf101

23. Tan W, Zhou L, Li X, Yang X, Chen Y, Yang J. Automated vessel segmentation in lung CT and CTA images via deep neural networks. *J X-Ray Sci Technol*. 2021;29(6):1123-1137. doi:10.3233/XST-210955

24. Liu J, Zhang Z, Niu B, et al. A Custom Annotated Dataset for Segmentation of Pulmonary Veins, Arteries, and Airways. *Sci Data*. 2025;12(1):1806. doi:10.1038/s41597-025-06074-6

25. Setio AAA, Traverso A, de Bel T, et al. Validation, comparison, and combination of algorithms for automatic detection of pulmonary nodules in computed tomography images: The LUNA16 challenge. *Med Image Anal*. 2017;42:1-13. doi:10.1016/j.media.2017.06.015

26. Isensee F, Jaeger PF, Kohl SAA, Petersen J, Maier-Hein KH. nnU-Net: a self-configuring method for deep learning-based biomedical image segmentation. *Nat Methods*. 2021;18(2):203-211. doi:10.1038/s41592-020-01008-z

27. Kirchhoff Y, Rokuss MR, Roy S, et al. Skeleton Recall Loss for Connectivity Conserving and Resource Efficient Segmentation of Thin Tubular Structures. *arXiv*. Preprint posted online July 17, 2024:arXiv:2404.03010. doi:10.48550/arXiv.2404.03010

28. Wasserthal J, Breit HC, Meyer MT, et al. TotalSegmentator: Robust Segmentation of 104 Anatomic Structures in CT Images. *Radiol Artif Intell*. 2023;5(5):e230024. doi:10.1148/ryai.230024

29. Książek J, Dziedzic R, Jelitto-Górska M, et al. Pomorski Pilotażowy Program Badań wczesnego wykrywania raka Płuca – doniesienie wstępne. *Ann Acad Medicae Gedanensis*. 2009;39:73-82.

30. Ostrowski M, Marjański T, Dziedzic R, et al. Ten years of experience in lung cancer screening in Gdańsk, Poland: a comparative study of the evaluation and surgical treatment of 14 200 participants of 2 lung cancer screening programmes. *Interact Cardiovasc Thorac Surg*. 2019;29(2):266-274. doi:10.1093/icvts/ivz079

31. Tushar FI, Wang A, Dahal L, et al. AI in Lung Health: Benchmarking Detection and Diagnostic Models Across Multiple CT Scan Datasets. *arXiv*. Preprint posted online October 29, 2025:arXiv:2405.04605. doi:10.48550/arXiv.2405.04605

32. Vegas-Sánchez-Ferrero G, Ledesma-Carbayo MJ, Washko GR, San José Estépar R. Harmonization of chest CT scans for different doses and reconstruction methods. *Med Phys*. 2019;46(7):3117-3132. doi:10.1002/mp.13578

33. Reynolds D. Gaussian Mixture Models. In: *Encyclopedia of Biometrics*. Springer, Boston, MA; 2009:659-663. doi:10.1007/978-0-387-73003-5_196

34. Lee TC, Kashyap RL, Chu CN. Building Skeleton Models via 3-D Medial Surface Axis Thinning Algorithms. *CVGIP Graph Models Image Process*. 1994;56(6):462-478. doi:10.1006/cgip.1994.1042

35. Silversmith W, Bae JA, Li PH, Wilson AM. seung-lab/kimimaro: Zenodo Release v1. Published online September 29, 2021. doi:10.5281/zenodo.5539913

36. Visvalingam M, Whyatt JD. Line generalisation by repeated elimination of points. *Cartogr J*. 1993;30(1):46-51. doi:10.1179/000870493786962263

37. Marsh BP, Chada N, Sanganna Gari RR, Sigdel KP, King GM. The Hessian Blob Algorithm: Precise Particle Detection in Atomic Force Microscopy Imagery. *Sci Rep*. 2018;8(1):978. doi:10.1038/s41598-018-19379-x

# Figure legends

**Figure 1.** Processing pipeline for bronchovascular bundle modelling.

**Figure 2.** Results of RONALD's bronchovascular bundle segmentation. The tracheobronchial walls are slightly lighter to accentuate the airflow. The marked malignant nodule is excluded from the structures.

**Figure 3.** Results of AirRC's bronchovascular bundle segmentation. The marked malignant nodule is inside the structures.

**Figure 4.** Results of TotalSegmentator's bronchovascular bundle segmentation. The marked malignant nodule is inside the structures.